\documentclass[12pt,twoside]{article}   %For printing on two sides of page
\usepackage[super,sort,comma]{natbib}
\usepackage{fancyhdr}		%Gives headers and footers defined below
\newcommand{\captionv}[3]{\begin{center}\parbox{#1cm}{\caption[#2]{{\sf #3}}}
        \end{center}}
\usepackage[section]{placeins}   %
\usepackage{graphicx}

\makeatletter \renewcommand\@biblabel[1]{$^{#1}$} \makeatother
\newcommand{\cen}[1]{\begin{center} #1 \end{center}}

\usepackage[mathlines]{lineno}
\usepackage{hyperref}
\hypersetup{ colorlinks,
    citecolor=blue,
    filecolor=blue,
    linkcolor=blue,
    urlcolor=blue
}

\usepackage{xcolor}
\definecolor{gray}{rgb}{0.6,0.6,0.6}
\definecolor{red}{rgb}{0.85,0,0}
\definecolor{green}{rgb}{0,0.85,0}
\definecolor{blue}{rgb}{0,0,0.85}
\definecolor{beige}{rgb}{0.92,0.87,0.78}
\usepackage[all]{hypcap}    %causes link to figures to go to figure, not caption
\newcommand{\argmin}{\mathop{\mathrm{argmin}}}
\newcommand{\norm}[1]{\Vert#1\Vert}
\newcommand{\xn}[1][n]{\mathbf{x}^\mathit{(#1)}}
\newcommand{\xnprev}[1][n]{\mathbf{x}^\mathit{(#1-\mathrm{1})}}

\usepackage{amsmath}
\newcommand{\hl}[1]{#1}

\begin{document}

\cen{\sf {\Large {\bfseries Computationally Efficient Deep Neural Network for Computed Tomography Image Reconstruction } \\  
\vspace*{10mm}
Dufan Wu\textsuperscript{a,b}, Kyungsang Kim\textsuperscript{a,b}, and Quanzheng Li\textsuperscript{a,b}} \\
\textsuperscript{a} Center for Advanced Medical Computing and Analysis, Massachusetts General Hospital and Harvard Medical School, Boston MA, US 02114 \\
\textsuperscript{b} Gordon Center for Medical Imaging, Massachusetts General Hospital and Harvard Medical School, Boston MA, US 02114
\vspace{5mm}\\
Version typeset \today\\
}

\pagenumbering{roman}
\setcounter{page}{1}
\pagestyle{plain}
Author to whom correspondence should be addressed. email: li.quanzheng@mgh.harvard.edu \\
% note, probably best not to use a student's e-mail as it won't be valid for
% very long.

\begin{abstract}
\noindent {\bf Purpose:} Deep-neural-network-based image reconstruction has demonstrated promising performance in medical imaging for under-sampled and low-dose scenarios. However, it requires large amount of memory and extensive time for the training. It is especially challenging to train the reconstruction networks for three-dimensional computed tomography (CT) because of the high resolution of CT images. The purpose of this work is to reduce the memory and time consumption of the training of the reconstruction networks for CT to make it practical for current hardware, while maintaining the quality of the reconstructed images. \\

{\bf Methods:} We unrolled the proximal gradient descent algorithm for iterative image reconstruction to finite iterations and replaced the terms related to the penalty function with trainable convolutional neural networks (CNN). The network was trained greedily iteration by iteration in the image-domain on patches, which requires reasonable amount of memory and time on mainstream graphics processing unit (GPU). To overcome the local-minimum problem caused by greedy learning, we used deep UNet as the CNN and incorporated separable quadratic surrogate with ordered subsets for data fidelity, so that the solution could escape from easy local minimums and achieve better image quality. \\

{\bf Results:} The proposed method achieved comparable image quality with state-of-the-art neural network for CT image reconstruction on 2D sparse-view and limited-angle problems on the low-dose CT challenge dataset. The difference in root-mean-square-error (RMSE) and structural similarity index (SSIM) was within $[-0.23, 0.47]$ HU and $[0, 0.001]$ respectively with 95\% confidence level. For 3D image reconstruction with ordinary-size CT volume, the proposed method only needed 2 GB graphics processing unit (GPU) memory and 0.45 seconds per training iteration as minimum requirement, whereas existing methods may require 417 GB and 31 minutes. The proposed method achieved improved performance compared to total-variation- and dictionary-learning-based iterative reconstruction for both 2D and 3D problems. \\

{\bf Conclusions:} We proposed a training-time computationally efficient neural network for CT image reconstruction. The proposed method achieved comparable image quality with state-of-the-art neural network for CT reconstruction, with significantly reduced memory and time requirement during training. The proposed method is applicable to 3D image reconstruction problems such as cone-beam CT and tomosynthesis on mainstream GPUs. \\

\end{abstract}
%\note{This is a sample note.}

\newpage     %may or may not be needed

%The table of contents is for drafting and refereeing purposes only. Note
%that all links to references, tables and figures can be clicked on and
%returned to calling point using cmd[ on a Mac using Preview or some
%equivalent on PCs (see View - go to on whatever reader).
%\tableofcontents

\newpage

\setlength{\baselineskip}{0.7cm}      %double spacing		

\pagenumbering{arabic}
\setcounter{page}{1}
\pagestyle{fancy}
\section{Introduction}
In computed tomography (CT), ill-posed image reconstruction problems are encountered in many applications. In breast tomosynthesis, the projections are only acquired in an angular range of approximately $50^\circ$\cite{wu2004comparison}, which is far less than the requirement for conventional analytical reconstruction algorithms ($180^\circ$ + fan angle). Limited-angular sampling is also used in mobile C-arm CT for intraoperative imaging, where the angular coverage is less than $180^\circ$\cite{nett2008tomosynthesis, schafer2011mobile}. Sparse-view sampling is encountered in many emerging applications, such as cardiac CT with high time resolution\cite{lauzier2012time} and CT with distributed sources\cite{wang2008outlook, cramer2018stationary}, where the number of projections per rotation is far less than that in current helical CT or cone-beam CT (CBCT). Under either limited-angle or sparse-view scenarios, images reconstructed by conventional filtered backprojection (FBP) will be accompanied with artifacts, and more advanced reconstruction methods are needed. 

Iterative image reconstruction with penalties have been studied for solving ill-posed reconstruction problems. The penalty functions are designed to exploit certain characteristics of medical images, such as smoothness, edge sparsity, low-rank, etc\cite{sidky2008image, elbakri2002statistical, kim2015sparse, kim2017low, xu2012low, Wu_2013}. In recent years, machine learning and deep neural networks have been applied to image reconstruction and achieved promising results. Deep neural networks incorporate complex nonlinear penalty functions\cite{goodfellow2016deep}, which better describe the characteristics of medical images compared to handcrafted penalties. The parameters of the networks can be decided efficiently from the training data, saving the enormous effort required for parameter tuning. 

Deep neural networks were initially introduced to CT image reconstruction as denoisers in the image domain, where it maps the FBP results from ill-posed data to better images\cite{jin2017deep, kang2017deep, chen2017low, wolterink2017generative, yang2018low}. However, the results could be sub-optimal because information may be inevitably lost during the initial FBP. To better incorporate sinogram into the network, unsupervised and semi-supervised learning have been investigated, where the networks are trained in the image domain but embedded into conventional iterative reconstruction\cite{wu2017iterative, gupta2018cnn, he2018optimizing}. Hyperparameters still need to be carefully tuned for both training and testing phase. 

Supervised learning usually has better performance compared to unsupervised and semi-supervised learning given same amount of data, because the neural networks are trained and tested to solve the same problem. In image reconstruction, supervised learning is achieved by "unrolled network", where the networks are built by unrolling iterative reconstruction algorithms and replacing terms relating to the penalty functions with trainable neural network components, such as convolutional layers and nonlinear activations. Several networks has been proposed for medical image reconstruction based on algorithms including gradient descent, primal-dual, alternating direction method of multiplier (ADMM), etc\cite{adler2018learned, chen2018learn, sun2016deep}. The modalities also vary in CT, magnetic resonance (MR), photoacoustic tomography, etc\cite{hammernik2018learning, schlemper2018deep, hauptmann2018model}.

Despite of the promising performance of unrolled networks for image reconstruction, the training is very computational expensive in terms of both memory and time. For efficient training of deep neural networks, some intermediate results have to be stored in the memory, which can be hundreds of times larger than the input image\cite{goodfellow2016deep}. The excessive memory cost becomes a challenge for current graphics processing units (GPU) for 3D CT images with relatively high resolution. For the widely used convolutional neural networks (CNN), this challenge can be overcome by training on local patches rather than the entire images\cite{zhang2017beyond}. However, for unrolled networks, patch-based training is not available due to the existence of system matrix in the network. As the consequence, the unrolled networks are not applicable to 3D image reconstruction problem such as tomosynthesis and conebeam CT due to the constrain of hardware. 

In this work we proposed a novel computationally efficient unrolled network for CT image reconstruction. The memory consumption problem was overcome by greedy iteration-wise training on patches in the image domain. Instead of training the entire network based on the final output, the proposed network was trained for each unrolled iteration sequentially. To enable patch-based training, the network was built from the proximal gradient descent algorithm\cite{parikh2014proximal} where the system matrix was decoupled from the proximal mapping to be trained. The proximal mapping was parameterized with deep U-Net\cite{ronneberger2015u} to mitigate the local-minimum problem induced by the greedy training. Separable quadratic surrogate with ordered subsets (OS-SQS)\cite{elbakri2002statistical} was also incorporated to further help the training escaping from local-minimum and achieve better image quality. 

The method was evaluated on the Low-dose CT Challenge\cite{aapm2017low} dataset and gave similar image quality to an existing unrolled network for 2D sparse-view and limited-angle reconstruction. The memory and time cost for 3D reconstruction was also estimated for both methods by extrapolation from small-sized problems. Furthermore, the proposed method was applied to 3D reconstruction problems and demonstrated better image quality than total variation and dictionary learning methods\cite{hou2018separable, xu2012low}. 

\section{Preliminaries}
In this section we will briefly introduce the unrolled network and explain why it is challenging to be trained for 3D image reconstruction problems. 

\subsection{Unrolled network}
Iterative image reconstruction for ill-posed problem solves the following unconstrained optimization problem:
\begin{linenomath*}
\begin{equation}\label{eq1}
\hl{\mathbf{x^*=\argmin_\mathbf{x}\norm{Ax-b}_w^2+{\beta}\mathit{R}(x)},}
\end{equation}
\end{linenomath*}
where \hl{$\mathbf{x^*}$} is the image to be reconstructed, $\mathbf{A}$ is the system matrix, $\mathbf{b}$ is the sinogram, $\mathbf{w}$ is the noise weighting, $R(\mathbf{x})$ is the penalty function, and $\beta$ is the hyperparameter to balance between the data fidelity and prior information. 

We will demonstrate the unrolled network based on proximal gradient descent\cite{parikh2014proximal} of Eq. \ref{eq1}:
\begin{linenomath*}
\begin{equation}\label{gd}
\xn=\textrm{prox}_{\gamma\beta R}\left\{ \mathbf{\xnprev-\gamma
A^\mathit{T}w(A\xnprev-b)} \right\},
\end{equation}
\end{linenomath*}
where $\mathbf{x}^{(n)}$ is the image at the $n$th iteration, and $\gamma$ is the step size. And the proximal operator $\textrm{prox}$ is defined as:
\begin{linenomath*}
\begin{equation}
\textrm{prox}_f(\mathbf{x})=\argmin_\mathbf{u}\left\{ f(\mathbf{u}) + \frac{1}{2}\norm{\mathbf{u-x}}_2^2 \right\}
\end{equation}
\end{linenomath*}

The unrolled network is built by replacing $\textrm{prox}_{\gamma\beta R}$ with CNNs and setting $\gamma$ as trainable variables in Eq. \ref{gd}:
\begin{linenomath*}
\begin{equation}\label{gd_network}
\xn=f_R \left\{ \mathbf{\xnprev-\gamma^{(\mathit{n})}
A^\mathit{T}w(A\xnprev-b) ; \Theta^\mathit{(n)}} \right \}, n=1,2,\dots,N,
\end{equation}
\end{linenomath*}
where $f_R(\mathbf{x};\mathbf{\Theta})$ is a CNN with input $\mathbf{x}$ and parameter $\mathbf{\Theta}$. $\gamma^{(n)}$ and $\mathbf{\Theta}^{(n)}$ are parameters to be trained, which are different across iteration $n$ for higher capacity of the network. $N$ is the number of iterations to truncate at. Better image quality can be achieved with larger $N$ at the cost of increased computational cost. 

The parameters are trained by solving:
\begin{linenomath*}
\begin{equation}\label{gd_train}
\hl{\gamma^{*(1)}, \mathbf{\Theta}^{*(1)}, \dots, \gamma^{*(N)}, \mathbf{\Theta}^{*(N)}=\argmin_{\gamma^{(1)}, \mathbf{\Theta}^{(1)}, \dots, \gamma^{(N)}, \mathbf{\Theta}^{(N)}} \sum_i{\norm{\mathbf{x}_i^{(N)} - \mathbf{x}_i^{ref}}_2^2},}
\end{equation}
\end{linenomath*}
where \hl{$\mathbf{x}_i^{(N)}$} and $\mathbf{x}_i^{ref}$ are the $i$th reconstructed image and the corresponding reference image respectively. \hl{$\mathbf{x}_i^{(N)}$ is determined recursively by the neural network \ref{gd_network}, which is a function of $\gamma^{(1)}, \mathbf{\Theta}^{(1)}, \dots, \gamma^{(N)}, \mathbf{\Theta}^{(N)}$.} The reference images are usually achieved from fully-sampled data. Stochastic gradient descent (SGD) algorithms are mostly used for solving Eq. \ref{gd_train} where the gradients are calculated via the chain rule of derivatives (backpropagation)\cite{goodfellow2016deep}. 

\subsection{Challenges for training}
During the training of neural networks, some intermediate results (featuremaps) need to be stored in memory for efficient backpropagation (see appendix A). In deep neural networks, the memory required to store the featuremaps is usually hundreds times the size of input images, which is very challenging for 3D CT images because of their high resolution. The large featuremaps also lead to slow backpropagation calculation. It will be demonstrated in section \ref{sec_cost} that it may require 417 GB GPU memory and 31 minutes per training iteration to train the learned primal-dual network\cite{adler2018learned} for $640\times 640\times 128$ CT images. On the contrary, existing mainstream GPUs only have less than 16 GB memory, which is severely insufficient for the training of unrolled networks. 

Patch-based training is an efficient way to reduce memory consumption to train 3D CNNs\cite{wolterink2017generative,zhang2017beyond}. For CNNs, each input pixel can only influence limited neighborhood in the output, so a compact support input will also give an output with compact support. Denote the CNN to be trained as $F(\mathbf{x}; \mathbf{\Theta})$, whole-image-based training such as Eq. \ref{gd_train} is equivalent to patch based training with proper patch sampling:
\begin{linenomath*}
\begin{equation}\label{train_patch_final}
\argmin_\mathbf{\Theta} \sum_i \norm {F(\mathbf{x}_i; \mathbf{\Theta}) - \mathbf{x}_i^{ref}}_2^2 = 
\argmin_\mathbf{\Theta} \sum_i \sum_k \norm {F(\mathbf{P}_{ik}\mathbf{x}_i; \mathbf{\Theta}) - \mathbf{E}_{ik}\mathbf{x}_i^{ref}}_2^2,
\end{equation}
\end{linenomath*}
where $\mathbf{P}_{ik}$, $\mathbf{E}_{ik}$ are the $k$th patch extraction matrix on the $i$th training images in the input and output space respectively. Since $\mathbf{P}_{ik}\mathbf{x}$ is much smaller, the corresponding featuremaps greatly reduced in size, and the training can be easily fit into current GPUs. 

However, for unrolled networks such as in Eq. \ref{gd_network}, any pixel in the input could affect the entire image through the system matrix $\mathbf{A}$. As the consequence, a compact support input will no longer give an output with compact support. Eq. \ref{train_patch_final} no longer holds unless both $\mathbf{P}_{ik}$ and $\mathbf{E}_{ik}$ cover the entire image. So the memory consumption problem persists due to the large size of input. More details are given in appendix B. 

\section{Methods}
\subsection{Network structure and training algorithm}
The key to reduce memory cost is to decouple system matrix from the neural network so that the network can be trained on patches. To achieve the decoupling, the training must be constrained in one unrolled iteration, which gave the greedy training: instead of training the network with Eq. \ref{gd_train}, the parameters were learned sequentially w.r.t. $n$:
\begin{linenomath*}
\begin{equation}\label{greedy_proto}
\hl{\gamma^{*(n)}, \mathbf{\Theta}^{*(n)} = \argmin_{\gamma^{(n)}, \mathbf{\Theta}^{(n)}} \sum_i \norm{ \mathbf{x}_i^{(n)} - \mathbf{x}_i^{ref} }_2^2, 
\textrm{ given } \gamma^{*(1)}, \mathbf{\Theta}^{*(1)}, \dots, \gamma^{*(n-1)}, \mathbf{\Theta}^{*(n-1)}}
\end{equation}
\end{linenomath*}

As demonstrated in Eq. \ref{gd_network}, the network $f_R(\mathbf{x,\Theta})$ is a CNN in the image domain, which could be efficiently trained on patches. By commencing the greedy training in Eq. \ref{greedy_proto}, the system matrix was decoupled from the CNN training because it involved only the input to the CNN and gradients did not backpropagate through it. 

It is known that greedy algorithm suffers the problem of local minimum and yield suboptimal solutions. The training at each $n$ mapped $\mathbf{x}_i^{(n-1)}$ to $\mathbf{x}_i^{ref}$ with networks of limited capacity, and $\mathbf{x}_i^{(n)}$ might finally stuck in a bad local minimum. To compensate for the bad local minimums, it was essential to use deep structure of CNN $f_R(\mathbf{x,\Theta})$ instead of the shallow networks used in existing unrolled networks\cite{adler2018learned, chen2018learn}. Deeper networks would help $\mathbf{x}_i^{(n)}$ escaping from easier local minimums and moving closer to the reference images. We employed UNet for its large capacity and receptive field. 

Another approach to mitigate the local-minimum problem was applying stronger perturbation to $\mathbf{x}^{(n-1)}$. The changes applied by the original gradient descent on $\mathbf{x}^{(n-1)}$ was relatively weak, especially if $\mathbf{x}$ was initialized from FBP. We used OS-SQS\cite{elbakri2002statistical} to acquire $\mathbf{y}^{(n-1)}$ as the perturbated $\mathbf{x}^{(n-1)}$:
\begin{linenomath*}
\begin{equation}\label{ossqs}
\begin{aligned}
&\mathbf{y}^{(n-1)}_m = \mathbf{y}^{(n-1)}_{m-1}-\frac{M\mathbf{A}_m^T\mathbf{w}_m(\mathbf{A}_m\mathbf{y}^{(n-1)}_{m-1} - \mathbf{b}_m)}{\mathbf{A}^T\mathbf{wA1}}
, m=1,2,\dots,M \\
&\mathbf{y}^{(n-1)}=\mathbf{y}^{(n-1)}_M, \mathbf{y}^{(n-1)}_0=\xnprev, 
\end{aligned}
\end{equation}
\end{linenomath*}
where $M$ is the number of subsets, $\mathbf{A}_m$, $\mathbf{w}_m$ and $\mathbf{b}_m$ are the system matrix, noise weighting, and sinogram corresponding to the $m$th subset. 

Both $\mathbf{x}^{(n-1)}$ and $\mathbf{y}^{(n-1)}$ were fed to the network as two channels and $\gamma^{(n)}$ was implicitly included in the network parameters. The final unrolled network became:
\begin{linenomath*}
\begin{equation}\label{network_structure}
\hl{\xn=f\left(\mathbf{\xnprev}, \mathbf{y}^{(n-1)}; \mathbf{\Theta}^{\mathit{(n)}} \right), n=1,2,\dots,N,}
\end{equation}
\end{linenomath*}

\begin{figure}[t]
   \begin{center}
   \includegraphics{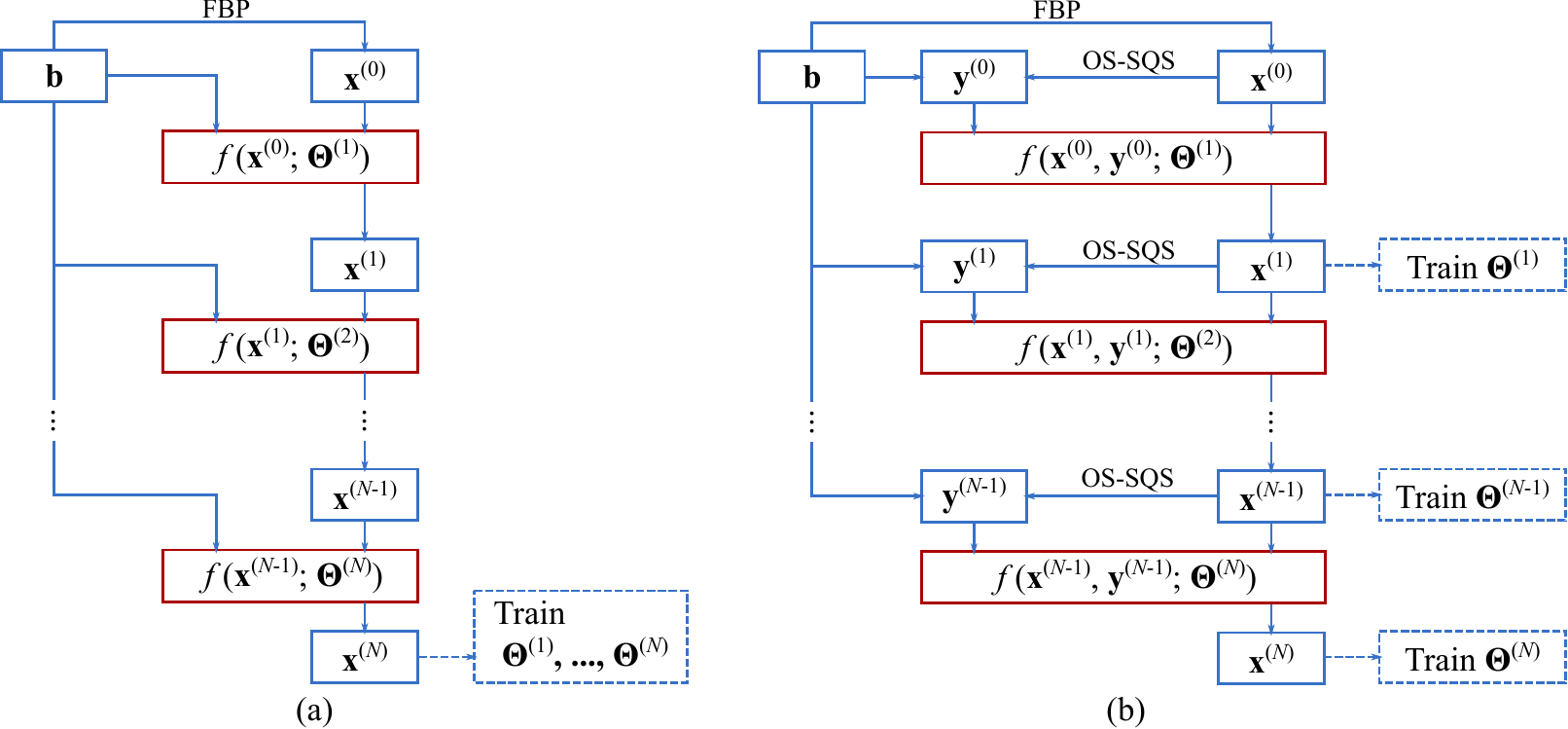}
   \captionv{12}{Network structures}
   {The outline of the unrolled networks and corresponding training schemes. (a) existing unrolled network; (b) the proposed method. 
   \label{fig_network} 
    }  %note label inside caption
    \end{center}
\end{figure}

Fig. \ref{fig_network} (b) demonstrated the structure of the proposed network and the corresponding training scheme. The structure of existing unrolled networks and their training schemes are given in Fig. \ref{fig_network} (a) for comparison. Fig. \ref{fig_network} demonstrated that existing unrolled networks learned all the parameters based on the final output $\mathbf{x}^{(N)}$, whereas the proposed method learned each set of parameters sequentially based on $\mathbf{x}^{(1)}, \mathbf{x}^{(2)}, \dots, \mathbf{x}^{(N)}$. 

\begin{figure}[t]
   \begin{center}
   \includegraphics{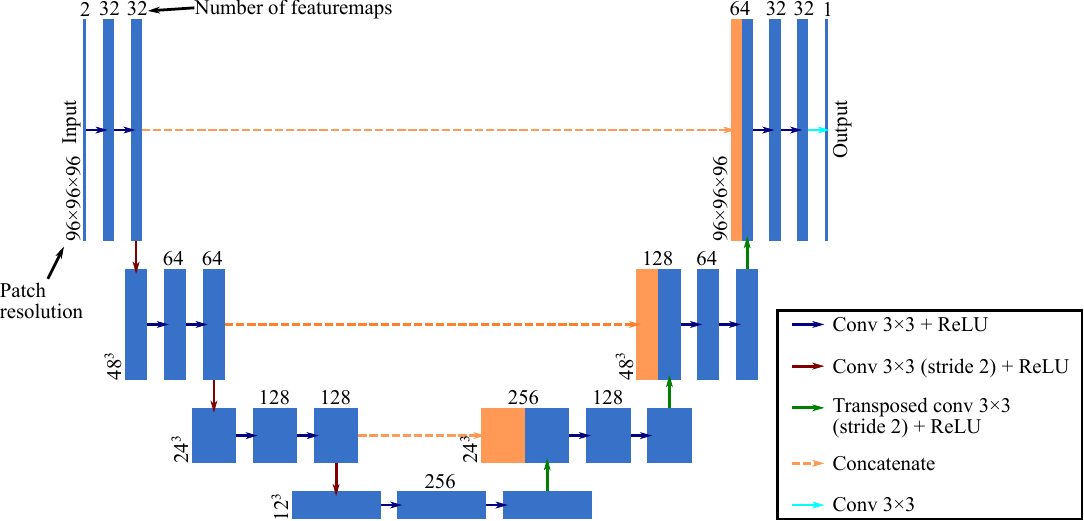}
   \captionv{12}{unet}
   {The structure of the UNet for $f(\mathbf{x}, \mathbf{y}; \mathbf{\Theta})$. 
   \label{fig_unet} 
    }  %note label inside caption
    \end{center}
\end{figure}

The UNet used to build $f(\mathbf{x}, \mathbf{y}; \mathbf{\Theta})$ is given in Fig. \ref{fig_unet}. The parameters in Fig. \ref{fig_unet} was training time parameters for 3D reconstruction. After the convolutional kernels were trained, it can be applied to larger patch resolutions during testing time because CNNs are shift invariant. We used $256\times256\times96$ patch size for 3D reconstruction during testing time. 

By adapting the greedy training prototype in Eq. \ref{greedy_proto} to the proposed network in Eq. \ref{network_structure} with patch-based training, the computationally efficient training algorithm for the proposed network is summarized in table \ref{training_algorithm}.

\begin{table}[t]
\begin{center}
\captionv{10}{Greedy training algorithm}{The greedy training algorithm for the proposed unrolled network
\label{training_algorithm}
\vspace*{2ex}
}
\begin{tabular} {p{0.9\linewidth}}
\hline
\\
\textbf{Algorithm 1}. Greedy training for the proposed unrolled network in Eq. \ref{network_structure} \\
\hline

\\
\hangindent=1.9 cm \textbf{INPUT}: Training sinograms $\mathbf{b}_i$, system matrices $\mathbf{A}_i$, noise weights $\mathbf{w}_i$, reference images $\mathbf{x}_i^{ref}$; 
Network structure $f(\mathbf{x}, \mathbf{y}; \mathbf{\Theta})$, number of unrolls $N$, number of subsets $M$;\\
\textbf{OUTPUT}: \hl{Trained network parameters $\mathbf{\Theta}^{*(1)}, \dots, \mathbf{\Theta}^{*(N)}$;}\\
\\
\textbf{1}. Initialize $\mathbf{x}_i^{(0)}$;\\
\textbf{2}. For $n=1,2,\dots,N$:\\
\textbf{3}. \hspace*{0.5cm} \textbf{OS-SQS}: Calculate $\mathbf{y}_i^{(n-1)}$ according to Eq. \ref{ossqs};\\
\textbf{4}. \hspace*{0.5cm} \textbf{Patch training}: \\
\hspace*{1.5cm} \hl{$\mathbf{\Theta}^{*(n)}=\argmin \sum_{i,k} \norm{f(\mathbf{P}_{ik}\mathbf{x}_i^{(n-1)}, \mathbf{P}_{ik}\mathbf{y}_i^{(n-1)}; \mathbf{\Theta}) - \mathbf{E}_{ik}\mathbf{x}_i^{ref}}_2^2$;}\\
\textbf{5}. \hspace*{0.5cm} \textbf{Prediction}: \hl{$\mathbf{x}_i^{(n)} = f(\mathbf{x}_i^{(n-1)}, \mathbf{y}_i^{(n-1)}; \mathbf{\Theta}^{*(n)})$;} \\
\textbf{6}. \hl{Return $\mathbf{\Theta}^{*(1)}, \dots, \mathbf{\Theta}^{*(N)}$.} \\
\\
\hline
\end{tabular}
\end{center}
\end{table}

In table \ref{training_algorithm}, step 3 to 5 completed the training of network at unrolled iteration $n$. Specifically, step 4 was the neural network training step. The training took patched input $\mathbf{P}_{ik}\mathbf{x}^{(n-1)}$ and $\mathbf{P}_{ik}\mathbf{y}^{(n-1)}$ instead of the whole images $\mathbf{x}^{(n-1)}$ and $\mathbf{y}^{(n-1)}$, and greatly reduced the memory and time consumption because of the reduced image size. 

\hl{A good property of the greedy training algorithm in table \ref{training_algorithm} was its monotony w.r.t. unroll number $n$:}
\begin{linenomath*}
\begin{equation}
\hl{\sum_i \norm{\mathbf{x}_i^{(n)}-\mathbf{x}_i^{ref}}_2^2 = 
\sum_i \norm{f(\mathbf{x}_i^{(n-1)}, \mathbf{y}_i^{(n-1)}; \mathbf{\Theta}^{*(n)})-\mathbf{x}_i^{ref}}_2^2 \leq
\sum_i \norm{\mathbf{x}_i^{(n-1)}-\mathbf{x}_i^{ref}}_2^2,}
\end{equation} 
\end{linenomath*}
\hl{where the inequality holds because that for the UNet given in figure \ref{fig_unet}, there exists $\mathbf{\Theta}'$ that $ \forall \mathbf{x} \textrm{ and } \mathbf{y}, f(\mathbf{x}, \mathbf{y}; \mathbf{\Theta}')=\mathbf{x}$. $\mathbf{\Theta}'$ could be constructed by setting certain convolutional kernels in the networks to identical filters.} This property saved the effort to tune unroll number $N$ during parameter tuning, where $N$ can be gradually increased until the performance saturated. 

\subsection{Computational cost analysis}
As the main purpose of the proposed network and training algorithm was to reduce the training-time computational cost of unrolled networks, it is necessary to analyze how much performance gain will be achieved by the proposed method in terms of memory and time. 

We assume that the training images had resolution $N_x \times N_y \times N_z$, projections had resolution $N_u \times N_v$ and there were $N_p$ views. GPU memory consumption were mainly contributed by the featuremaps of CNNs. 

For unrolled network in Eq. \ref{gd_network}, assuming that $f_R(\mathbf{x}; \mathbf{\Theta})$ had $N_{f_R}$ convolutional layers with $N_c$ featuremaps per layer, then an $N$-unroll network would require $O(N_xN_yN_zN_{f_R}N_cN)$ memory for training. For the greedy training algorithm, assuming that patch size was $N_{p_x} \times N_{p_y} \times N_{p_z}$ and network had $N_f$ convolutional layers with $N_c$ featuremaps per layer, then the memory consumption was $O(N_{p_x}N_{p_y}N_{p_z}N_fN_c)$. The GPU memory cost of the proposed method over unrolled network training would be
\begin{linenomath*}
\begin{equation}\label{memory_complexity}
O\left(  \frac{N_{p_x}N_{p_y}N_{p_z}}{N_xN_yN_z} \times \frac{N_f}{N_{f_R}N}\right)
\end{equation}
\end{linenomath*}

Although the greedy training algorithm required deeper networks, it did not expand through unrolls and usually it would hold that $N_f \leq N_{f_R}N$. The largest saving came from the first term which related to patch-based training, where $N_xN_yN_z$ could be more than 50 times $N_{p_x}N_{p_y}N_{p_z}$. Furthermore, the memory cost for the greedy training was determined by the patch size rather than the image size, meaning that it can be applied to high-resolution data without increasing memory cost. 

For patch-based training, each iteration may take more than one patches for more stable training (minibatch), but it is very straight forward to calculate the gradient of the network patch by patch and accumulate them for the final update. Hence, we did not include the minibatch size in the memory complexity in Eq. \ref{memory_complexity}.

The time cost for training follows approximately the same complexity, since the forward and backward propagation of CNN have linear complexity w.r.t. image size and number of layers. The complexity w.r.t. $N_c$ became $O(N_c^2)$ but it was the same for both methods. In addition to CNN training, unrolled network's time complexity also included the $N$ calculations of $\mathbf{A}$ and $\mathbf{A}^T$, which was $O(N(N_x+N_y+N_z)N_uN_vN_p)$. However, such calculation can be neglected for greedy training since they only need to be calculated once per unroll instead of every training iteration. For the greedy training, batch size $N_b$ and number of unrolls $N$ should be included in the time complexity analysis. The training time cost for the proposed method over unrolled network was:
\begin{linenomath*}
\begin{equation}
O\left(  \frac{N_{p_x}N_{p_y}N_{p_z}}{N_xN_yN_z} \times \frac{N_f}{N_{f_R}} \times N_b \right) + 
O\left(  \frac{N_{p_x}N_{p_y}N_{p_z}}{(N_x + N_y+ N_z)N_uN_vN_p} \times N_f \times N_b \right)
\end{equation}
\end{linenomath*}

The saving on training time by algorithm 1 was not as much as its saving on GPU memory due to the additional factors $N_bN$, but it is still considerable because of the large ratio of $N_xN_yN_z$ to $N_{p_x}N_{p_y}N_{p_z}$.

\section{Experimental setups}
\subsection{Datasets}
The training and testing data were from the Low-dose CT Challenge\cite{aapm2017low}, which consisted of the sinograms and images from 10 abdomen scans. 8 patients were randomly selected for training whereas the rest 2 were used for testing in our experiments. The sinograms were rebinned to multi-slice fanbeam before down-sampling and reconstruction\cite{Noo_1999}, where all the geometric parameters after rebinning were kept the same as the original ones, including views per rotation, source to center distance, source to detector distance, and detector resolution. The slice thickness after rebinning was 1 mm. The axial resolution was $640 \times 640$, and the axial pixel sizes were the same with the images reconstructed by the scanners, which varied from $0.66\times 0.66$ to $0.8 \times 0.8$ mm\textsuperscript{2}.

Ill-posed image reconstruction problems were generated by down-sampling the rebinned sinograms. 2D studies were used to compare the proposed method with existing unrolled networks. For 2D studies, we used $4\times$, $8\times$ and $12\times$ sparse-view and $180^{\circ}$ and $150^{\circ}$ limited-angle projections. We also further used 3 mm slice thickness instead of 1 mm to increase signal-to-noise ratio (SNR). 50 slices were randomly selected for both training and testing. For the 3D studies, we demonstrated the feasibility of the proposed method in 3D on $4\times$ sparse-view and $180^{\circ}$ limited-angle problems. 

\hl{Besides real data experiments, noiseless simulation studies were also done using forwarded projected sinograms from the fully-sampled FBP results. The forward projection geometry was the same with the rebinned geometry. Compared to real data studies, the noiseless simulation provided ground truth for more accurate evaluation of the algorithms with quantitative metrics such as root mean square error (RMSE) and structural similarity index (SSIM)\cite{wang2004image}. 2D studies with $18\times$, $36\times$ sparse-view and $180^{\circ}$, $150^{\circ}$ limited-angle were used in the simulation. Compared to real experiments, a larger down-sampling rate could be used in noiseless sparse-view studies because the reconstructed image quality was no longer limited by total photon flux. }

\subsection{Parameters}
For the proposed method, we used unroll number $N=10$ and subsets number $M=32$. The subsets for OS-SQS were determined by reversing the bits order of the projections' serial number. The network $f(\mathbf{x}, \mathbf{y};\mathbf{\Theta})$ was a UNet with depth of 4, whose detailed parameters are shown in Fig. \ref{fig_unet}. For 2D studies, we used $96 \times 96$ patch size with minibatch size of 40, and 40 patches were randomly extracted from each training slice. During testing time, the whole slice of $640\times 640$ was fed to the trained UNet. For 3D studies, we used $96 \times 96 \times 96$ patch size with minibatch size of 2, and 100 patches were randomly extracted from each patients. The patch size during testing time was $256 \times 256 \times 96$, with step size of $192 \times 192 \times 72$. 100 epochs of Adam algorithm\cite{kingma2014adam} with $10^{-4}$ learning rate were used for 2D training whereas 250 epochs were used for 3D training. The training patches were randomly flipped at all directions as data augmentation. The pixel values were normalized to $\mathrm{HU} / 1000$ before feeding to the networks. 

There were 3 most important hyperparameters for the proposed network: number of unrolls $N$, number of subsets $M$, and the complexity of $f(\mathbf{x}, \mathbf{y};\mathbf{\Theta})$. Their influence was investigated on 2D $150^{\circ}$ limited angle reconstruction problems by changing one of the parameters while fixing the other two. We used the depth of UNet for the complexity of $f(\mathbf{x}, \mathbf{y};\mathbf{\Theta})$ in the hyperparameter study. 

We implemented the learned primal-dual network as the reference method of unrolled networks. It was chosen for its relatively compact size and high degree of freedom. \hl{Its structure and parameters were kept the same with that in the work by Adler and Oktem\cite{adler2018learned}}, where 10 unrolls were used and a 3-layer CNN was applied in both image and projection domain at each unroll. The primal-dual networks were trained for 500 epochs with $10^{-3}$ learning rate with cosine annealing. The pixel values were normalized to $(\mathrm{HU} + 1000) / 1000$ before being fed to the networks. 

Total variation (TV) minimization and dictionary learning were also implemented as reference methods of iterative algorithms. \hl{OS-SQS algorithm was used for TV reconstruction\cite{hou2018separable}, where 16 subsets were used with Nesterov's accelerated gradient\cite{devolder2014first}. For dictionary learning, dictionaries were learned by K-SVD\cite{aharon2006k} from the 8 training patients on $8 \times 8$ patches with 256 atoms and sparse level of 10, where the parameters were taken from the work by Xu et al.\cite{xu2012low}} During reconstruction, overlapped dictionary patches were sampled at step size of 6 with random perturbation to reduce computational cost. The reconstruction algorithm was alternative optimization with OS-SQS with 16 subsets. For 3D reconstruction, 3 dictionaries were applied to axial, sagittal and coronal planes rather than a 3D dictionary to save computational time\cite{wu2017iterative}. \hl{Both TV and dictionary learning were implemented on GPU. For real data studies, 100 iterations were used for TV and 25 iterations were used for dictionary learning. For noiseless simulations, 200 iterations were used for TV and 100 iterations were used for dictionary learning. The iteration number was selected to achieve near converged results under realistic computational time. The hyperparameter $\beta$ for both algorithms were selected by parameter sweeping and selecting the ones with best RMSE. }

All the reconstructions were initialized from FBP with Hann filter. Riess weighting\cite{schafer2017modified, riess2013tv} was used for the limited angle reconstructions. Reference images for both training and testing were reconstructed by FBP with Hann filter from fully-sampled projections.

\subsection{Image quality evaluation}
\hl{We calculated the RMSE and SSIM for the testing results as the qualitative metrics to evaluate the algorithms' performance.} The SSIMs were calculated within the liver window, which was [-160, 240] HU. For 3D reconstructions, the SSIMs were calculated by averaging the SSIMs in of each slice. 
To evaluate the image quality of learned primal-dual and the proposed method, confidence intervals (CIs) were calculated for the difference of the two method:
\begin{linenomath*}
\begin{equation}\label{dif_var}
D=X_\textrm{primal-dual}-X_\textrm{proposed},
\end{equation}
\end{linenomath*}
where $X_\textrm{primal-dual}$ and $X_\textrm{proposed}$ are the metrics (RMSE or SSIM) for the images reconstructed by the learned primal-dual and the proposed network respectively. CIs were calculated for the aggregated results from all 5 sampling conditions, as well as individual sampling conditions.  

\subsection{Computational cost estimation}\label{comput_cost_method}
We realized the neural network with Tensorflow 1.11\cite{abadi2016tensorflow} and reconstruction with CUDA 9.2 on a GTX 1080 Ti, whose available GPU memory was approximately 11 GB. $\mathbf{A}$ and $\mathbf{A}^T$ were calculated with Siddon's fast ray-tracing\cite{siddon1985fast} on the fly and did not require extra memory. 

Since learned primal-dual required too much GPU memory to train on 3D images with practical size, we evaluated its memory consumption on small-scale problems first, then extrapolated it to practical scale with multilinear model:
\begin{linenomath*}
\begin{equation}\label{memory_cost}
\textrm{memory cost}=\sum_{i=0}^{1}\sum_{j=0}^{1}\sum_{k=0}^{1}c_{ijk}N^iN_p^jN_z^k,
\end{equation}
\end{linenomath*}
where $N$ is the number of unrolls, $N_p$ is the number of views, $N_z$ is the number of slices, and $c_{ijk}$ are the coefficients to be fitted. Due to caches needed for neural network training, we also included terms with $i$, $j$ or $k$ equals to 0. 

To evaluate the minimum required GPU memory for a configuration $(N, N_p, N_z)$, we tried different maximum allowed GPU memory in Tensorflow with binary search to find the minimum memory that would not cause overflow. We sampled $N$ from 1 to 10, $N_p$ between 64 and 192, and $N_z$ between 5 and 20. All the sample points with minimum required memory less than the physical memory of our GPU were used to fit the model in Eq. \ref{memory_cost}.

The time cost were estimated with the same multilinear extrapolating model. We also modeled the memory cost and time per iteration to train the proposed network w.r.t. size of minibatches with linear model. Larger minibatch requires more GPU memory and time per iteration, but leads to more stable training\cite{goodfellow2016deep}.

\section{Results}
\hl{We will first demonstrate the noiseless simulation results, followed by 2D and 3D real data studies.} Then the computational cost of the learned primal-dual and the proposed method will be shown. At last, we will discuss the influence of hyperparameters on the proposed method. 

\subsection{\hl{Noiseless simulation}}

\begin{figure}[t]
   \begin{center}
   \includegraphics{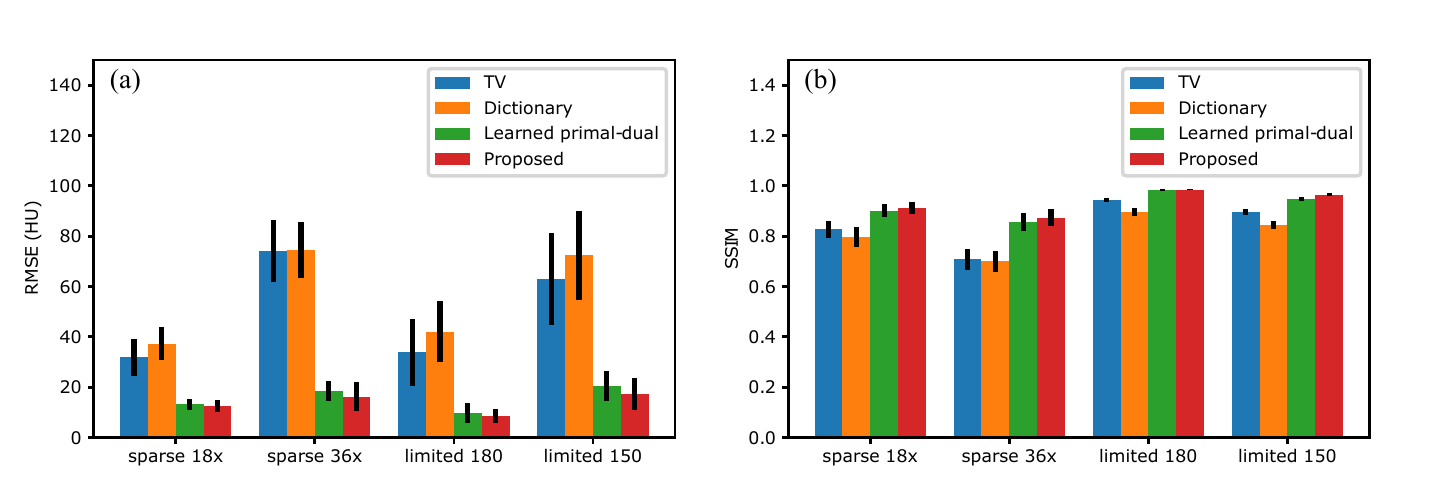}
   \captionv{12}{Image metrics simulation}
   {\hl{The mean and standard deviations of RMSEs and SSIMs of the testing results compared to reference images in noiseless simulation. (a) RMSEs; (b) SSIMs. The error bars are the standard deviation. }
   \label{fig_img_quality_simul} 
    }  %note label inside caption
    \end{center}
\end{figure}

\begin{figure}[t]
   \begin{center}
   \includegraphics{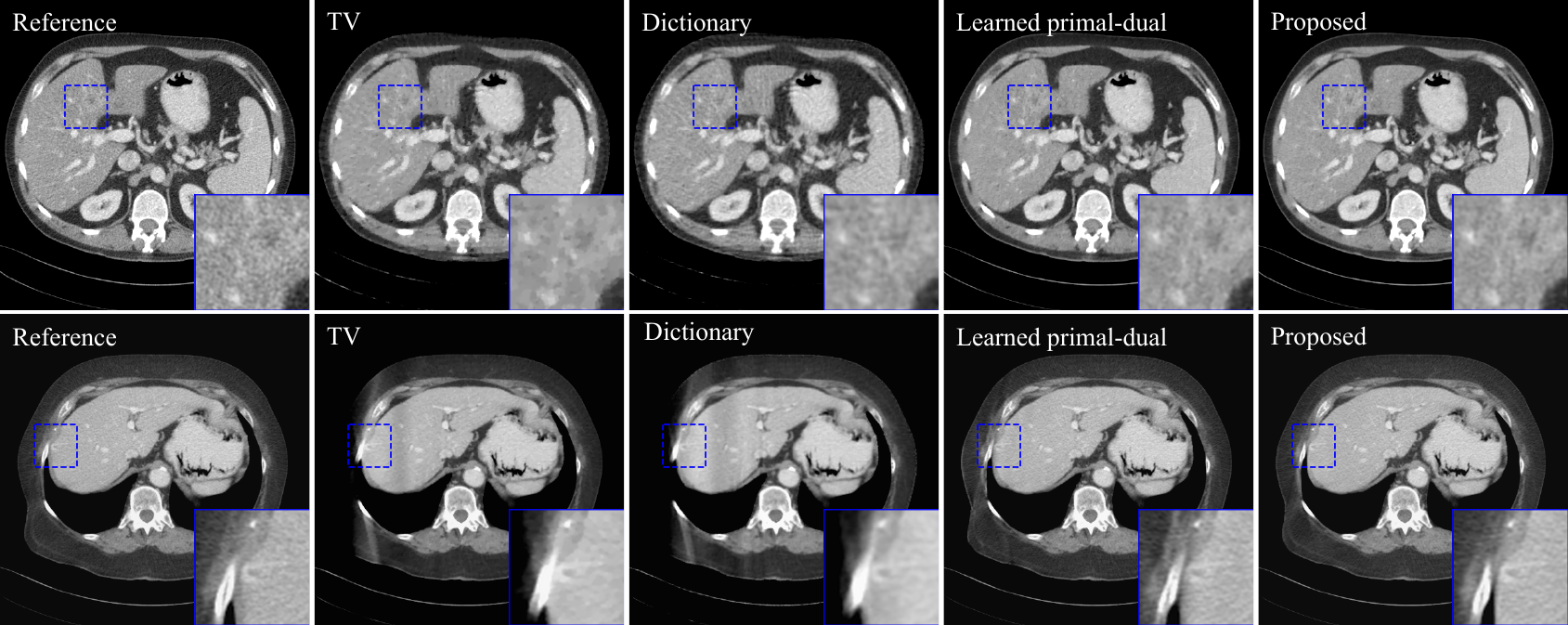}
   \captionv{12}{Images simulation}
   {\hl{Testing results in the noiseless simulation. First row is $18\times$ sparse-view study and second row is $180^\circ$ limited-angle study. Lesions are zoomed in. The display windows are [-160, 240] HU. }
      \label{fig_simul} 
    }  %note label inside caption
    \end{center}
\end{figure}

\hl{The methods were trained and tested for $18\times$, $36\times$ sparse-view and $180^{\circ}$, $150^\circ$ limited-angle simulated noiseless projections. RMSEs and SSIMs were calculated for each testing slice compared to the reference images. The mean and standard deviation for each configuration are shown in Fig. \ref{fig_img_quality_simul}. Some slices from the testing results are given in Fig. \ref{fig_simul}.}

\hl{Consistent improvement was observed for deep-neural-network-based method over TV and dictionary learning, and the reconstructed image quality was close between learned primal-dual and the proposed method. In Fig. \ref{fig_simul}, network-based results achieved better texture preservation and less shading artifacts compared to TV and dictionary learning. There was no significant visual difference on the images reconstructed by learned primal-dual and the proposed method. }

\subsection{Reconstructed image quality: 2D studies}

\begin{figure}[t]
   \begin{center}
   \includegraphics{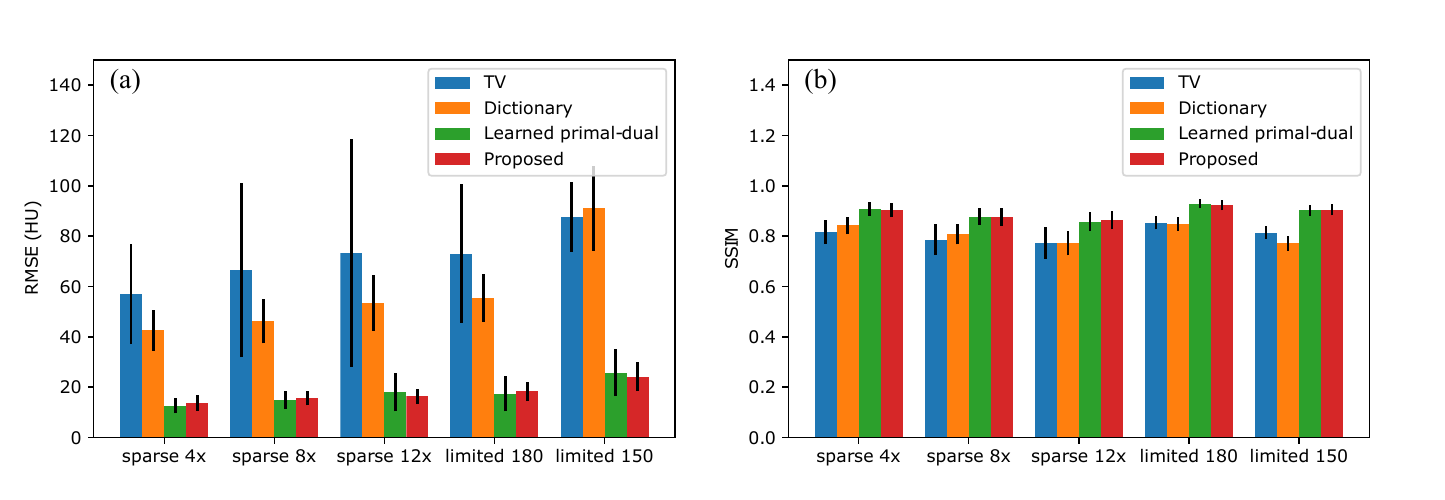}
   \captionv{12}{Image metrics}
   {The mean and standard deviations of RMSEs and SSIMs of the testing results \hl{in 2D real data studies}. (a) RMSEs; (b) SSIMs. The error bars are the standard deviation. 
   \label{fig_img_quality} 
    }  %note label inside caption
    \end{center}
\end{figure}

\begin{figure}[t]
   \begin{center}
   \includegraphics{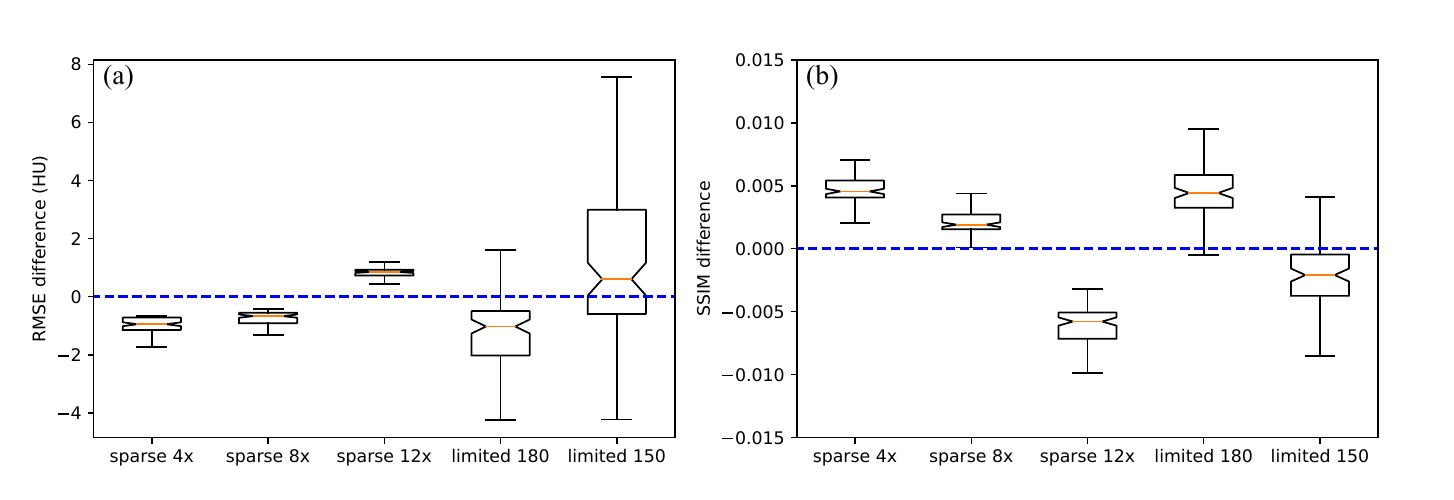}
   \captionv{12}{Image cmp}
   {The box and whisker plot of the performance difference between learned primal-dual and the proposed network \hl{in 2D real data studies}: (a) RMSE of learned primal-dual minus RMSE of the proposed network; (b) SSIM of learned primal-dual minus SSIM of the proposed network. The 0-errors were marked with dashed blue lines. Outliers are not plotted. 
      \label{fig_img_cmp} 
    }  %note label inside caption
    \end{center}
\end{figure}

We trained and tested the methods for $4\times$, $8\times$, $12\times$ sparse-view and $180^{\circ}$, $150^\circ$ limited-angle projections. RMSEs and SSIMs were calculated for each testing slice compared to the reference images and their mean and standard deviation for each configuration are shown in Fig. \ref{fig_img_quality}. The RMSEs and SSIMs of learned primal-dual minus that of the proposed method are plotted as box and whisker plots in Fig. \ref{fig_img_cmp}. The 95\% CIs were calculated under three scenarios: aggregation of all the sampling conditions, $180^\circ$ limited angle where learned primal-dual outperformed the proposed method the most, and sparse $12\times$ where the proposed method outperformed learned primal-dual the most. The 95\% CIs are given in table \ref{table_ci}. 

Fig. \ref{fig_img_quality} demonstrated significant advantage of the deep-neural-network-based methods over TV and dictionary learning w.r.t. RMSEs and SSIMs. The performance of the learned primal-dual and the proposed method were close to each other under both metrics. A more detailed comparison are given in Fig. \ref{fig_img_cmp}, which showed the majority of errors fell within 2 HU for RMSE and 0.01 for SSIM. Further analysis showed that learned primal-dual had slightly better performance for easier problems such as sparse $4\times$ and limited $180^{\circ}$. However, the 95\% CIs of SSIM for these cases were around 0.005, which meant no significant visual differences on the reconstructed images.

\begin{table}[t]
\begin{center}
\captionv{10}{CIs}{The 95\% confidence intervals of differences of RMSEs and SSIMs \hl{of the proposed method and learned primal-dual in 2D real data studies}
\label{table_ci}
\vspace*{2ex}
}
\begin{tabular} {p{4cm}p{4cm}p{4cm}}
\hline
\\
Scenarios & RMSE (HU) & SSIM \\
\hline
\\
Overall & $[-0.23, 0.47]$ & $[8.8\times 10^{-5}, 0.0010]$\\
\\
limited $180^{\circ}$ (Worst) & $[-1.76, 0.078]$ & $[0.0041, 0.0056]$\\
\\
sparse $12\times$ (Best) & $[0.51, 2.61]$ & $[-0.0077, -0.0060]$\\
\hline
\end{tabular}
\end{center}
\end{table}

\begin{figure}[t]
   \begin{center}
   \includegraphics{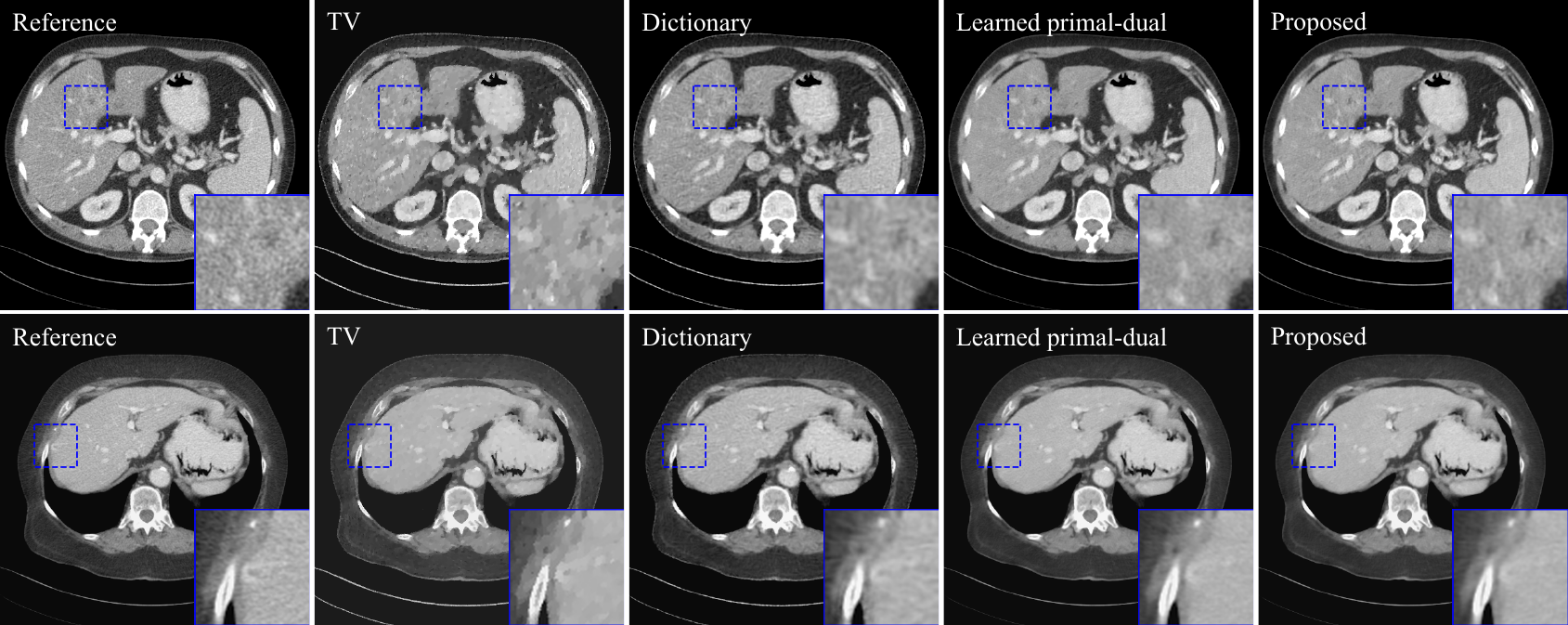}
   \captionv{12}{sparse2d}
   {\hl{The testing slices from 2D sparse $4\times$ study. Lesions are zoomed in. The gray scale windows are $[-160, 240]$ HU.}
      \label{fig_sparse2d} 
    }  %note label inside caption
    \end{center}
\end{figure}

\begin{figure}[t]
   \begin{center}
   \includegraphics{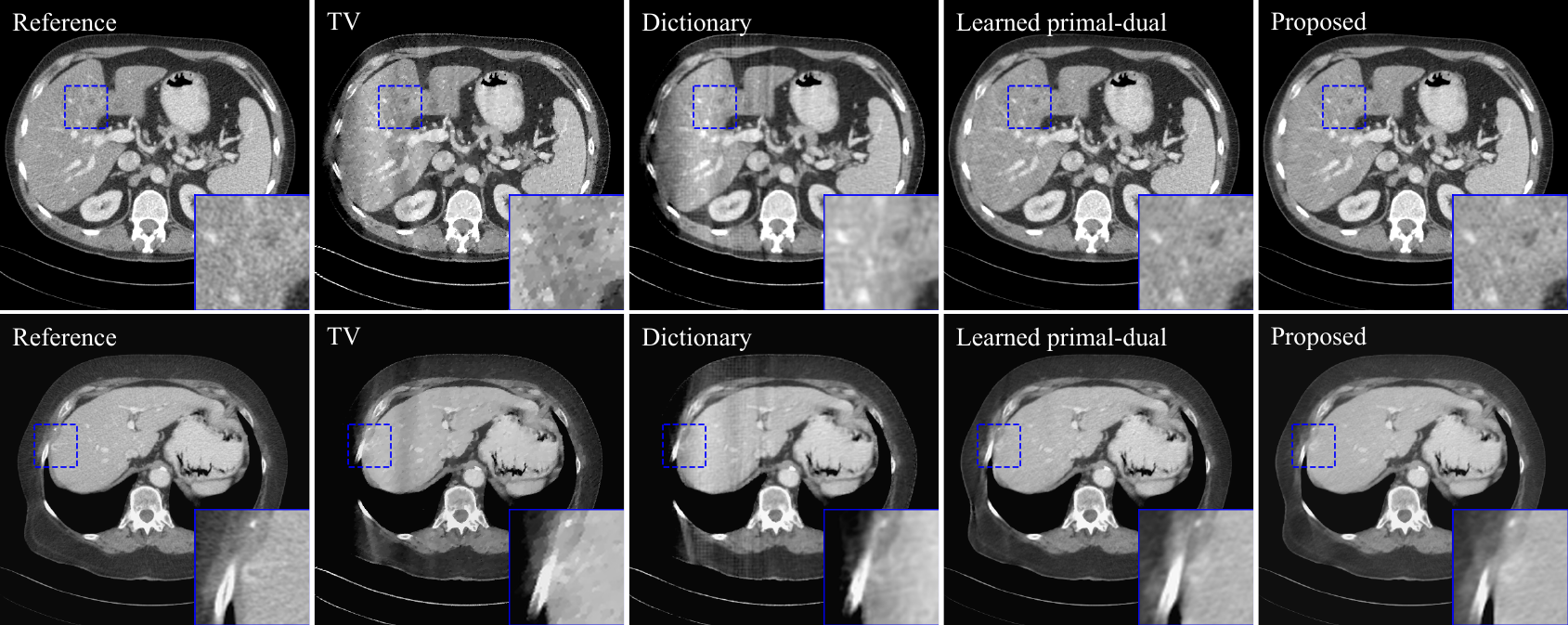}
   \captionv{12}{fig_limited2d}
   {\hl{The testing slices from 2D limited $180^\circ$ study. Lesions are zoomed in. The gray scale windows are $[-160,240]$ HU.}
      \label{fig_limited2d} 
    }  %note label inside caption
    \end{center}
\end{figure}

Some of the testing slices from the 2D sparse $4\times$ and limited $180^\circ$ studies are given in Fig. \ref{fig_sparse2d} and \ref{fig_limited2d}. The two network-based methods demonstrated less blurred structures compared to TV and dictionary learning results for the sparse view study. For the limited angle study, the network-based methods recovered the shading artifacts due to missing angles compared to TV and dictionary learning. The textures of the network-based methods were also closer to the reference images compared to TV and dictionary learning. There were no significant difference on the images reconstructed by the network-based methods, which was consistent with our conclusion drawn from analysis on RMSEs and SSIMs. Both methods generated results close to the reference images with slight loss of textures due to the L2 norm used in the network training. 

The zoom-ins in Fig. \ref{fig_sparse2d} and \ref{fig_limited2d} demonstrated the visibility of lesions in the reconstructed images. The lesion visibility in the results of learned primal-dual and proposed network had slightly worse contrast compared to the reference images, but the lesions were still visible. 

\subsection{Reconstructed image quality: 3D }
\begin{figure}[t]
   \begin{center}
   \includegraphics{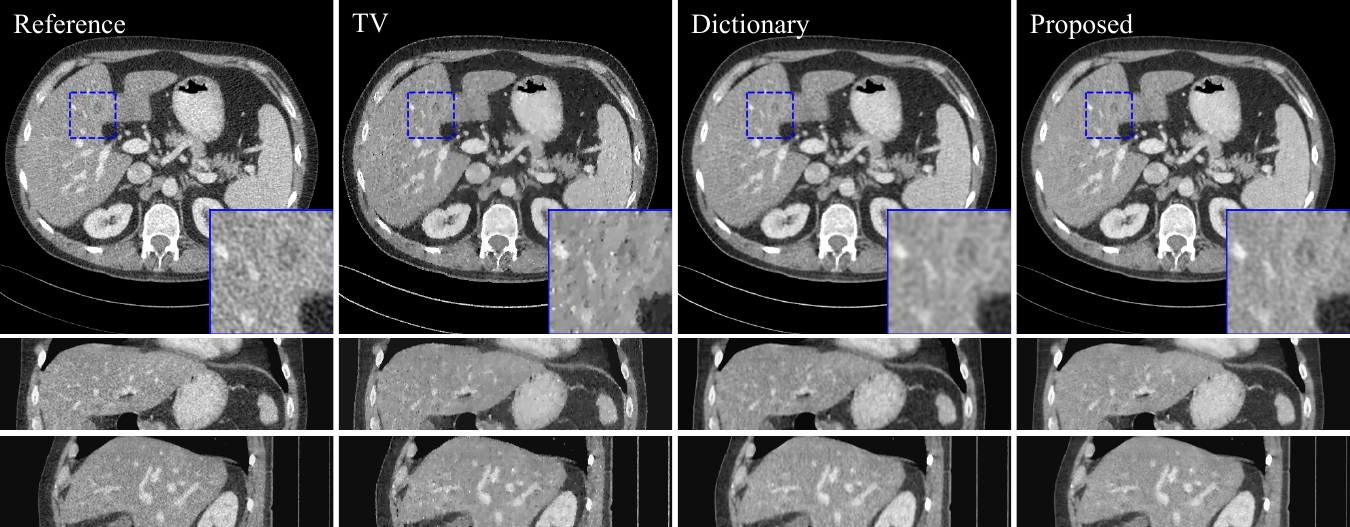}
   \captionv{12}{sparse3d}
   {\hl{The testing slices from 3D sparse $4\times$ study. Lesions are zoomed in in the axial views and marked with arrows in the sagittal and coronal views. The gray scale windows are $[-160, 240]$ HU.}
      \label{fig_sparse3d} 
    }  %note label inside caption
    \end{center}
\end{figure}

\begin{figure}[t]
   \begin{center}
   \includegraphics{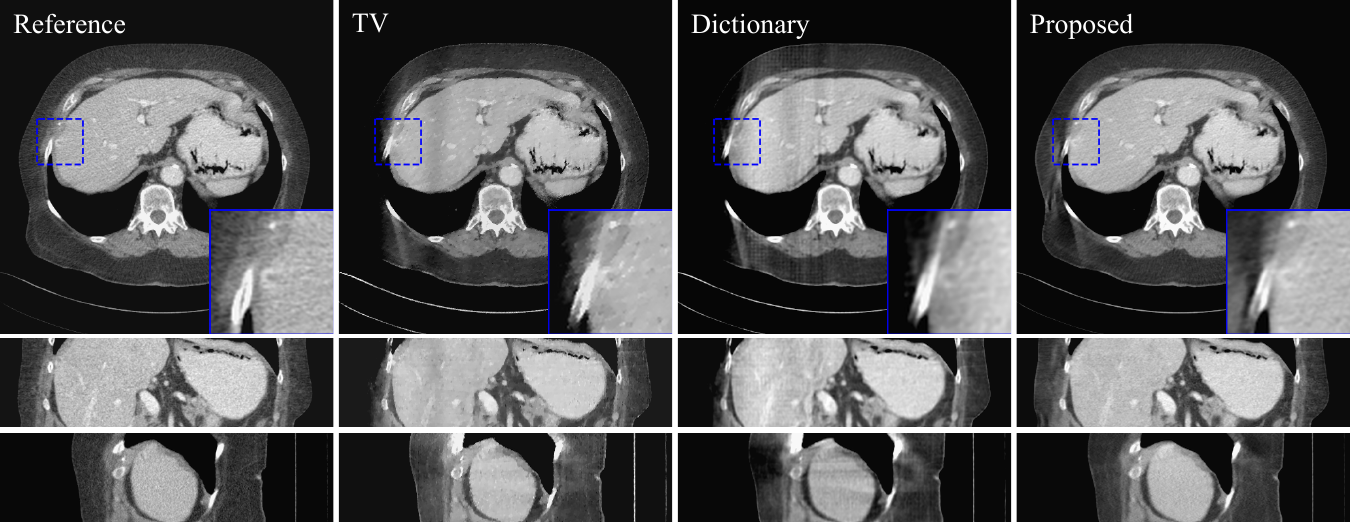}
   \captionv{12}{limited3d}
   {\hl{The testing slices from 3D limited $180^\circ$ study. Lesions are zoomed in in the axial views and marked with arrows in the sagittal and coronal views. The gray scale windows are $[-160,240]$ HU.}
      \label{fig_limited3d} 
    }  %note label inside caption
    \end{center}
\end{figure}

The 3D reconstructed images are given in Fig. \ref{fig_sparse3d} and \ref{fig_limited3d} for sparse $4 \times$ and limited $180^\circ$ studies. The RMSEs and SSIMs of the testing images and are given in table \ref{table_3d}. As stated before, learned primal-dual required too much GPU memory for practical-scaled problem and the training could not be done due to memory overflow. As the consequence, there was no results for learned primal-dual in 3D. 

The proposed network achieved the best performance on RMSEs and SSIMs over TV and dictionary learning for both 3D sparse-view and limited-angle problems. For sparse-view reconstruction, the results from the proposed method had texture closer to the reference images compared to TV and there was less blurring compared to dictionary learning. For limited-angle reconstruction, the proposed method significantly reduced the shading artifacts caused by the insufficient sampling angle. 

The lesions were also zoomed-in in the axial-views for the 3D results. The visibility of the small lesion in Fig. \ref{fig_limited3d} was improved compared to the corresponding 2D results in Fig. \ref{fig_limited2d}, since a larger neighbourhood was included in the 3D neural network. 

\begin{table}[t]
\begin{center}
\captionv{10}{RMSEs and SSIMs 3D}{The RMSEs and SSIMs of 3D reconstruction results.
\label{table_3d}
\vspace*{2ex}
}
\begin{tabular} {lllllll}
\hline
\\
Method && \multicolumn{2}{c}{sparse $4\times$} && \multicolumn{2}{c}{limited $180^\circ$} \\
\cline{3-4}\cline{6-7}
             && RMSE(HU) & SSIM && RMSE(HU) & SSIM \\
\hline
\\
TV && 50.13 & 0.817 && 65.56 & 0.849 \\
\\
Dictionary && 40.61 & 0.834 && 59.76 & 0.842\\
\\
Proposed && \textbf{22.24} & \textbf{0.878} && \textbf{26.88} & \textbf{0.902}\\
\hline
\end{tabular}
\end{center}
\end{table}

\subsection{Computational cost}\label{sec_cost}
The computational cost was estimated according to the approach described in section \ref{comput_cost_method}. Since we could not directly estimate the memory and time cost to train learned primal-dual for practical-sized problems due to GPU memory overflow, it was extrapolated from the estimation on small-sized problems with multilinear model. 

\begin{figure}[t]
   \begin{center}
   \includegraphics{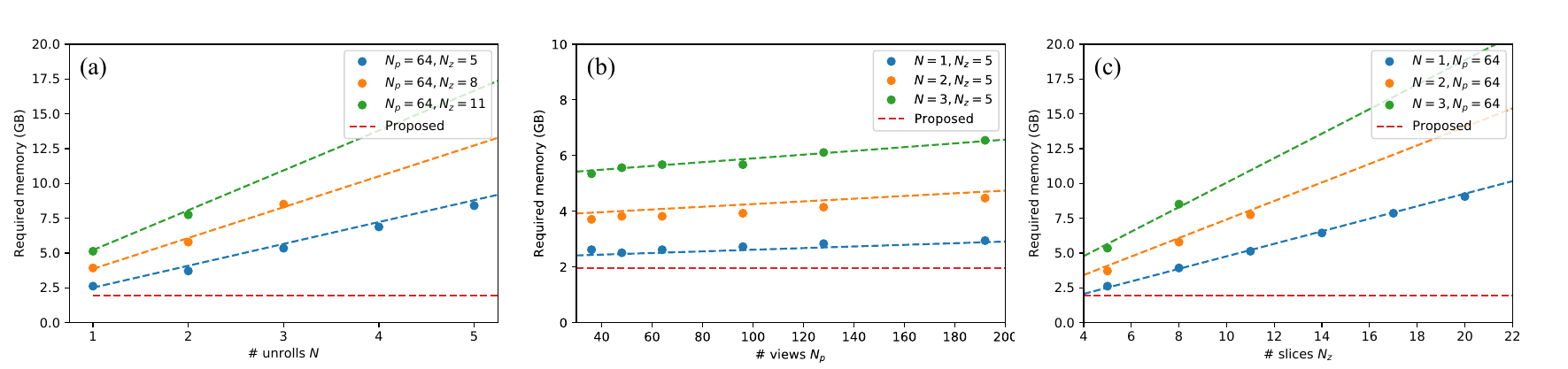}
   \captionv{12}{memory_curve}
   {The required GPU memory to train learned primal-dual with various unrolls $N$, views $N_p$ and slices $N_z$. The dots are measured points and the dashed lines are model prediction. The dashed red line is the memory cost of the proposed method. 
      \label{fig_memory_curve} 
    }  %note label inside caption
    \end{center}
\end{figure}

We fitted the multilinear model in Eq. \ref{memory_cost} with available $(N, N_p, N_z)$ for learned primal-dual and achieved coefficient of determination $r^2=0.9928$ for memory cost and $r^2=0.9989$ for time per iteration, suggesting good linear relationship. Some of the data points with fitted lines are given in Fig. \ref{fig_memory_curve}. The minimum required memory for the proposed method on $96\times 96\times 96$ patch is given as the dashed red line. 

\begin{figure}[t]
   \begin{center}
   \includegraphics{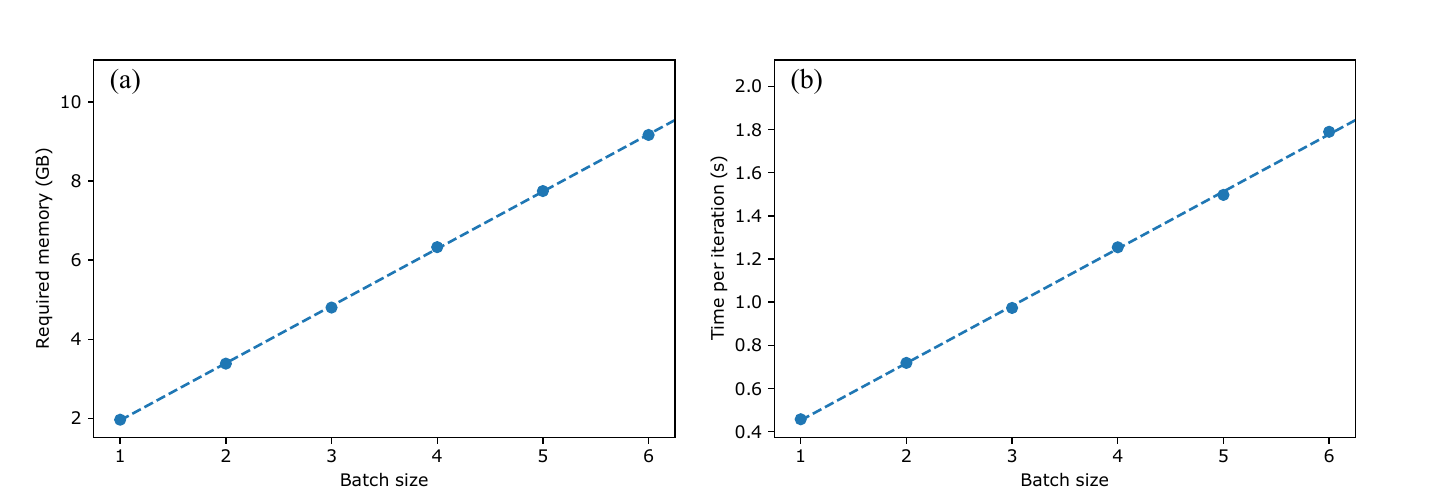}
   \captionv{12}{Memory and time proposed}
   {The required GPU memory and time per iteration to train the proposed network with different minibatch size. The dots are the measured points and the dashed lines are model prediction. The patch size was $96\times 96\times 96$.
      \label{fig_memory_proposed} 
    }  %note label inside caption
    \end{center}
\end{figure}

The memory cost and time per iteration for the proposed network w.r.t. different minibatch sizes are given in Fig. \ref{fig_memory_proposed}. The coefficient of determination was $r^2=0.9998$ for memory cost and $r^2=0.9995$ for time per iteration, suggesting good linear relationship. 

Fig. \ref{fig_memory_curve} demonstrated that the memory consumption increases quickly with $N$ and $N_z$ and easily went beyond the 11 GB memory of the GPU. We extrapolated the model to practical scale where the image size was $640\times 640\times 128$, the projection size was $768\times 576\times 128$, and the unroll number was 10. The required memory and time per training iteration were approximately 417 GB and 31 minutes respectively, which were beyond the reach of current mainstream hardware. The minimum requirement for the proposed method was 1.96 GB and 0.45 seconds with minibatch size of 1. For the minibatch size of 2 used in the 3D studies, the requirement was 3.38 GB and 0.71 seconds. Both memory cost and time per iteration were practical for current hardware.

\subsection{Influence of hyperparameters}\label{sec_hyper}
\begin{figure}[t]
   \begin{center}
   \includegraphics{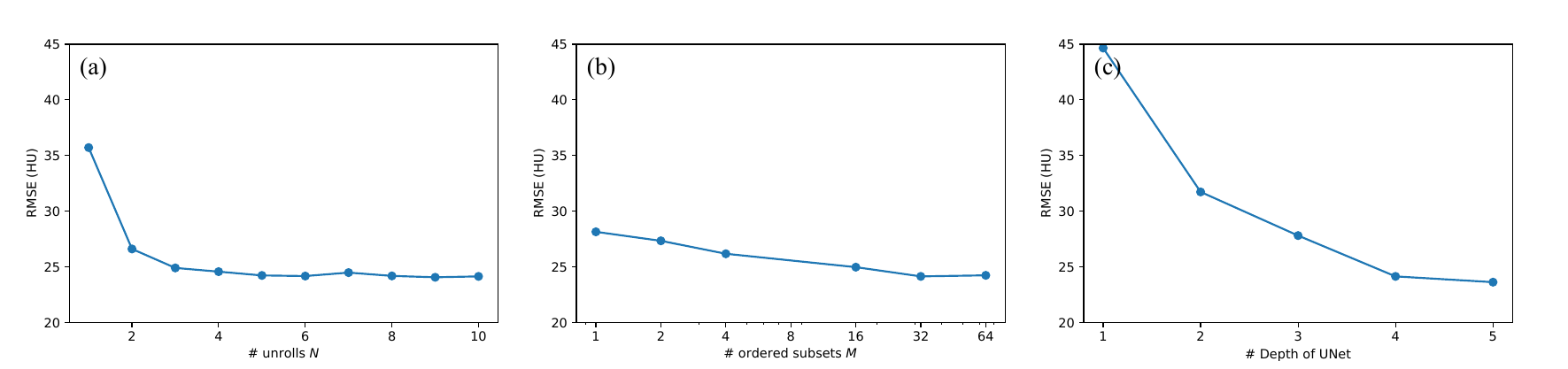}
   \captionv{12}{Hyperparameters}
   {The RMSEs of the testing images with different unroll number $N$, OS number $M$ and depth of UNet. The baseline configuration was $N=10$, $M=32$, $\textrm{depth}=4$. The problem was 2D $150^\circ$ limited angle reconstruction. 
      \label{fig_hyperparameters} 
    }  %note label inside caption
    \end{center}
\end{figure}

The influence of the 3 most important hyperparameters, unroll number $N$, OS number $M$ and depth of UNet at testing time are given in Fig. \ref{fig_hyperparameters}. The baseline configurations were $N=10$, $M=32$ and $\textrm{depth}=4$ for the three curves. Fig. \ref{fig_hyperparameters} was generated by alternating one of the hyperparameters while keeping the rest two at the baseline.

The depth of UNet had the largest influence on the image quality, which is in accordance with our previous assumption that deeper $f(\mathbf{x,y;\Theta})$ would help the greedy training find better local minimum. The unrolls number $N$ had moderate influence especially within the first 4 unrolls. The unroll curve was basically monotonic until convergence, where fluctuations were caused by the randomness in the initialization of neural networks and SGD algorithms. The OS number $M$ had the weakest influence compared to the other two factors, where the RMSE reduced by 4 HU when $M$ increased from 1 to 32. The relative weak correlation indicated that there was no need for careful tuning of $M$. 

\begin{figure}[t]
   \begin{center}
   \includegraphics{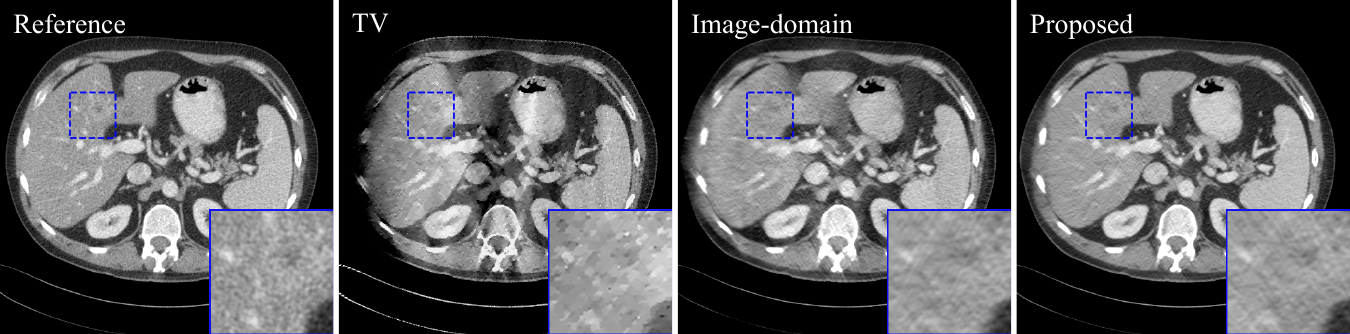}
   \captionv{12}{unroll}
   {\hl{The reconstructed images of 2D $150^\circ$ limited angle study. Image-domain result was the reconstructed images at unroll 1. TV result is provided to demonstrate the severity of limited angle artifacts. The display window is [-160, 240] HU.}
      \label{fig_unroll} 
    }  %note label inside caption
    \end{center}
\end{figure}

\hl{The unroll number $N$ versus RMSE curve also demonstrated the advantage of the proposed method over image-domain methods\cite{jin2017deep, kang2017deep, chen2017low, wolterink2017generative, yang2018low}. Each unroll in the proposed method composed of a deep UNet, which was widely used in networks-based image processing. The output at unroll 1, $\mathbf{x}^{(1)} = f(\mathbf{x}^{(0)}, \mathbf{y}^{(0)}; \mathbf{\Theta}^{*(1)})$ could be considered as approximated results from image-domain deep UNet, because $\mathbf{x}^{(0)}$ was FBP results and $\mathbf{y}^{(0)}$, which was the image after one iteration of OS-SQS, was close to $\mathbf{x}^{(0)}$. $\mathbf{x}^{(1)}$ and $\mathbf{x}^{(10)}$ of a testing slice in the 2D $150^\circ$ study are given in Fig. \ref{fig_unroll}. $\mathbf{x}^{(10)}$ had significantly reduced blurring and shading artifacts compared to $\mathbf{x}^{(1)}$, which indicated the effectiveness of the proposed method compared to image-domain UNet. }

\section{Conclusion and discussion}
In this paper we proposed a deep-neural-network-based image reconstruction method for ill-posed CT imaging problems. A neural network was built based on proximal gradient descent algorithm and trained with greedy algorithm. The greedy algorithm decoupled network training from system matrix, and it could be executed on local patches with greatly reduced memory and time cost compared to existing unrolled networks. The proposed method can be applied to 3D CT reconstruction, whose computational cost was too high for the training of existing network-based reconstruction methods. 2D experimental results demonstrated comparable image quality of the proposed method and the learned primal-dual on sparse-view and limited-angle problems. \hl{Besides real data studies, a noiseless simulation was also done for performance evaluation without interference from the noise in the reference images. The simulation results were consistent with that in the real data study, which justified the use of fully-sampled FBP as reference images for the calculation of RMSEs and SSIMs in real data studies.} The feasibility of the proposed method in 3D was also demonstrated, with better image quality compared to TV and dictionary learning. 

There were a few hyperparameters for the proposed methods as shown in section \ref{sec_hyper}. However, the effort required for hyperparameter tuning was not much increased compared to unrolled network. The network structure (depth of UNet) is almost the only part that needs tuning. Although the number of unrolls $N$ had large impact on image quality, the loss is monotonic to $N$ unless over-fitting happens, which requires reduction of complexity of $f(\mathbf{x,y;\Theta})$. The number of OS $M$ had relatively weak influence on the performance, it could be kept with around 30 projections per subsets during tuning of the network structure, then be fine-tuned afterwards. 

There was slight loss of textures in the images reconstructed by both network-based methods compared to the reference images. This was mainly due to the L2-norm used during the training in Eq. \ref{gd_train} and \ref{train_patch_final}. It has been demonstrated that replacing the L2-norm with adversarial loss\cite{goodfellow2014generative, wolterink2017generative} could greatly improve the textures. The adversarial loss can be easily incorporated into the proposed network by including it in the training of the last unroll $f(\mathbf{x,y;\Theta^{(\mathit{N})}})$. It is known that training with adversarial loss was unstable and may need careful parameter tuning\cite{arjovsky2017wasserstein}, incorporating it only in the last unroll would save effort for hyperparameter tuning. 

In this work we demonstrated the method on rebinned helical data to study its property and performance. Data rebinning is possible in helical data, which enabled our 2D studies and comparison with learned primal-dual. However, under many situations 3D to 2D data rebinning is not possible, such as cone-beam CT and tomosynthesis. The method can also be applied to other modalities, including 3D MR, positron emission tomography, etc.  

\newpage     %This is set up for tables and figures at end of text but they 
	     %may be in the text and is strongly preferred that way by some
	     %referees.

\clearpage

% following only if there is an appendix
\section*{Appendix}\label{appendix}
\addcontentsline{toc}{section}{\numberline{}Appendix}

\subsection*{A. Explaination on memory cost for CNN training}\label{explain_cnn}
The memory consumption during CNN training is further explained here. A CNN $F(\mathbf{x};\mathbf{\Theta})$ consists of sequential convolution and nonlinear operations. An $L$-layer plain CNN can be written as:
\begin{linenomath*}
\begin{equation}
F(\mathbf{x};\mathbf{\Theta}) = g^{(L)}\circ g^{(L-1)}\circ\dots\circ g^{(1)}(\mathbf{x})
\end{equation}
\end{linenomath*}

Let
\begin{linenomath*}
\begin{equation}
\mathbf{g}^{(l)}=g^{(l)}\circ g^{(l-1)}\circ\dots\circ g^{(1)}(\mathbf{x}), l=1,2,\dots,L,
\end{equation}
\end{linenomath*}
where $\mathbf{g}^{(l)}$ is called "featuremaps". The convolutional layers $g^{(l)}(\cdot)$ can be written as:
\begin{linenomath*}
\begin{equation}\label{plain_cnn_layer}
g^{(l)}(\mathbf{g}^{(l-1)})=\sigma^{(l)}\mathbf{(W^\mathit{(l)}*g^\mathit{(l-\mathrm{1})} + b^\mathit{(l)})}, l=1,2,\dots,L
\end{equation}
\end{linenomath*}
where $\sigma^{(l)}(\cdot)$ is a point-wise nonlinear function such as ReLU or PReLU\citep{nair2010rectified,he2015delving}, except that $\sigma^{(L)}(\cdot)$ in the last layer is identity mapping. $\mathbf{W}^{(l)}$ and $\mathbf{b}^{(l)}$ are the convolutional kernel and bias to be trained for the $l$th layer. 

For 3D images, the convolution is actually carried out in 4D space where the extra dimension is called "feature space". Let $\mathbf{y}^{(l)}=\mathbf{W}^{(l)} * \mathbf{g}^{(l-1)}$, we have:
\begin{linenomath*}
\begin{equation}
y_{j_1,j_2,j_3,c}^{(l)}=\sum_{c'=1}^{C^{(l-1)}}\sum_{j_1',j_2',j_3'}{W^{(l)}_{j_1',j_2',j_3',c}g_{j_1-j_1', j_2-j_2', j_3-j_3', c'}^{(l-1)}},
c=1,2,\dots,C^{(l)},
\end{equation}
\end{linenomath*}
where $j_1$, $j_2$, $j_3$ are the image space coordinates and $c$ is the feature space coordinate. $C^{(l)}$ is the length of the feature space in layer $l$. Since the input and output of $F(\mathbf{x};\mathbf{\Theta})$ are images, we have $C^{(0)}=C^{(L)}=1$.

Training of the CNNs needs derivatives of the loss in Eq. \ref{gd_train} w.r.t. variables such as $\mathbf{W}^{(l)}$. Denote the loss as $z$, according to chain rule of derivatives, we have:
\begin{linenomath*}
\begin{equation}\label{kernel_deriv}
\frac{\partial z}{\partial W^{(l)}_{j_1',j_2',j_3',c}}=\sum_{j_1,j_2,j_3,c}\frac{\partial z}{\partial y^{(l)}_{j_1,j_2,j_3,c}}
\frac{\partial y^{(l)}_{j_1,j_2,j_3,c}}{\partial W^{(l)}_{j_1',j_2',j_3',c}}
= \sum_{j_1,j_2,j_3,c'}\frac{\partial z}{\partial y^{(l)}_{j_1,j_2,j_3,c}}
g_{j_1-j_1', j_2-j_2', j_3-j_3', c'}^{(l-1)}
\end{equation}
\end{linenomath*}
\begin{linenomath*}
\begin{equation}\label{x_deriv}
\frac{\partial z}{\partial g_{j_1', j_2', j_3', c'}^{(l-1)}}=\sum_{j_1,j_2,j_3,c}\frac{\partial z}{\partial y^{(l)}_{j_1, j_2, j_3, c}}
\frac{\partial y^{(l)}_{j_1,j_2,j_3,c}}{\partial g_{j_1', j_2', j_3', c'}^{(l-1)}}
= \sum_{j_1,j_2,j_3,c}\frac{\partial z}{\partial y^{(l)}_{j_1, j_2, j_3, c}}
W^{(l)}_{j_1-j_1', j_2-j_2', j_3-j_3', c}
\end{equation}
\end{linenomath*}

Equation \ref{kernel_deriv} demonstrated that the featuremaps $\mathbf{g}^{(l)}$ need to be known to calculate the derivatives. Eq. \ref{x_deriv} indicated that the chain rule of derivatives must be applied from the end of the network (backpropagation). However, the featuremaps can only be calculated from the beginning of the network (forward propagation). Hence, for efficient training of the network, all the featuremaps must be stored in memory before the backpropagation starts, otherwise each $\mathbf{g}^{(l)}$ needs to be calculated from the beginning of the network whenever it is required. 

\subsection*{B. Patch-based training}\label{patch_training}
Plain CNN layers in Eq. \ref{plain_cnn_layer} have finite receptive field, which means that it maps any compactly supported input $\mathbf{g}^{(l-1)}$ to compactly supported output $\mathbf{g}^{(l)}$. Hence, the entire CNN $F(\mathbf{x; \Theta})$ also has finite receptive field. More complex CNNs, such as UNet, also have finite receptive field because additional layers (strided convolution and concatenating) also have finite receptive field. The finite receptive field means that for every compactly supported patching matrix $\mathbf{P}$ in the input domain, there exists a corresponding compactly supported matrix $\mathbf{E}$ in the output domain such that:
\begin{linenomath*}
\begin{equation}\label{train_patch_1}
F(\mathbf{P}\mathbf{x}; \mathbf{\Theta}) = \mathbf{E}F(\mathbf{x}; \mathbf{\Theta})
\end{equation}
\end{linenomath*}

Patch-based training is defined as:
\begin{linenomath*}
\begin{equation}
\mathbf{\Theta} = \argmin_\mathbf{\Theta} \sum_i \sum_k \norm {F(\mathbf{P}_{ik}\mathbf{x}_i; \mathbf{\Theta}) - \mathbf{E}_{ik}\mathbf{x}_i^{ref}}_2^2
\end{equation}
\end{linenomath*}

Due to Eq. \ref{train_patch_1}, we have:
\begin{linenomath*}
\begin{equation}\label{train_patch_2}
\begin{split}
&\sum_i \sum_k \norm {F(\mathbf{P}_{ik}\mathbf{x}_i; \mathbf{\Theta}) - \mathbf{E}_{ik}\mathbf{x}_i^{ref}}_2^2 \\
= &\sum_i \sum_k \norm {\mathbf{E}_{ik}F(\mathbf{x}_i; \mathbf{\Theta}) - \mathbf{E}_{ik}\mathbf{x}_i^{ref}}_2^2 \\
= &\sum_i \left( F(\mathbf{x}_i; \mathbf{\Theta}) - \mathbf{x}_i^{ref} \right)^T \left( \sum_k \mathbf{E}_{ik}^T\mathbf{E}_{ik} \right) \left( F(\mathbf{x}_i; \mathbf{\Theta}) - \mathbf{x}_i^{ref} \right)
\end{split}
\end{equation}
\end{linenomath*}

If the patching matrix $\mathbf{E}_{ik}$ is all-ones on its compact support, and samples uniformly over the images, we have
\begin{linenomath*}
\begin{equation}
\sum_k \mathbf{E}_{ik}^T\mathbf{E}_{ik} = C\cdot\mathbf{1},
\end{equation}
\end{linenomath*}
where $C$ is a constant. So we have, 
\begin{linenomath*}
\begin{equation}\label{train_patch_3}
\argmin_\mathbf{\Theta} \sum_i \sum_k \norm {\mathbf{E}_{ik}F(\mathbf{x}_i; \mathbf{\Theta}) - \mathbf{E}_{ik}\mathbf{x}_i^{ref}}_2^2 = 
\argmin_\mathbf{\Theta} \sum_i \norm {F(\mathbf{x}_i; \mathbf{\Theta}) - \mathbf{x}_i^{ref}}_2^2,
\end{equation}
\end{linenomath*}
which concludes the equivalency between the patch-based training and the original loss function as shown in Eq. \ref{train_patch_final}.

However, in unrolled networks $\mathbf{A}^T\mathbf{wAx}$ no longer has compact support even if $\mathbf{x}$ is compactly supported. Both $\mathbf{E}$ and $\mathbf{P}$ has to cover the entire image for Eq. \ref{train_patch_1} to hold. As the consequence, patch-based training is not available for unrolled networks. 

\section*{References}
\addcontentsline{toc}{section}{\numberline{}References}

% Following assumes you are using bibtex. However, for submission to the
% journal you MUST explicitly INCLUDE THE REFERENCES IN THE TEX FILE. 
% In that case you need the following

% The following is when using bibtex and picks up the example.bib file

%\bibliography{Explicit address of .bib file}
%\bibliography{./ComputEffReconNet_dufan}      %example.bib is on the same directory
% above points to where we find the master reference list
% and also causes the bibliography to be printed

% When creating your bibliography you should run bibtex on your local
% computer after running pdflatex on your .tex file. bibtex will
% generate a .bbl file.
% Copy the contents of this .bbl file into your main latex document,
% replacing the "\bibliography" command which was pointing at your .bib file.

% following defines style of .bbl file 

%\bibliographystyle{explicit relative path to medphy.bst}
\bibliographystyle{./medphy}    %if this is installed on your system,
				    %it is not essential to have the    ./

% Note that you need to typeset once, then run bibtex, then typeset another
% two times to get the references working properly.

\end{document}